\RequirePackage{fix-cm}
\documentclass[smallextended]{svjour3}       % onecolumn (second format)
\smartqed  % flush right qed marks, e.g. at end of proof

\usepackage[T1]{fontenc}%https://www.overleaf.com/project/6138b1e84a82b0cf1a9ecf59
\usepackage[utf8]{inputenc}
\usepackage{graphicx}
\usepackage{newtxtext,newtxmath} % Replaces mathptmx below, https://tex.stackexchange.com/questions/308145/class-svjour3-package-mathptmx-warning-there-are-no-bold-math-fonts-on-input-li
\usepackage{latexsym}
\usepackage{natbib}
\usepackage{url}
\usepackage{subcaption}
\usepackage{tikz}
\begin{document}

\title{Finnish Parliament ASR corpus
%\thanks{}
}
% Grants or other notes about the article that should go on the front
% page should be placed within the \thanks{} command in the title
% (and the %-sign in front of \thanks{} should be deleted)
%
% General acknowledgments should be placed at the end of the article.

\subtitle{Analysis, benchmarks and statistics}


\author{Anja Virkkunen*         \and
        Aku Rouhe               \and
        Nhan Phan               \and
        Mikko Kurimo
}

\authorrunning{Virkkunen et al.} % if too long for running head

\institute{*Corresponding author \\
           \at Department of Signal Processing and Acoustics, Aalto University, Espoo, Finland \\
           \email{firstname.lastname@aalto.fi}
}
           %A. Virkkunen \at
           %   Aalto University \\
%          %    Tel.: +123-45-678910\\
           %   \email{anja.virkkunen@aalto.fi}           %  \\
%          %   \emph{Present address:} of F. Author  %  if needed
           %\and
           %A. Rouhe \at
           %   Aalto University \\
           %   \email{aku.rouhe@aalto.fi}           %  \\
           %\and
           %N. Phan \at
           %   Aalto University \\
           %   \email{nhan.phan@aalto.fi}
           % \and
           %M. Kurimo \at
           %   Aalto University \\
           %   \email{mikko.kurimo@aalto.fi}           %  \\


\maketitle

\begin{abstract}
Public sources like parliament meeting recordings and transcripts provide ever-growing material for the training and evaluation of automatic speech recognition (ASR) systems. In this paper, we publish and analyse the Finnish parliament ASR corpus, the largest publicly available collection of manually transcribed speech data for Finnish with over 3000 hours of speech and 449 speakers for which it provides rich demographic metadata. This corpus builds on earlier initial work, and as a result the corpus has a natural split into two training subsets from two periods of time. Similarly, there are two official, corrected test sets covering different times, setting an ASR task with longitudinal distribution-shift characteristics. An official development set is also provided. We develop a complete Kaldi-based data preparation pipeline, and hidden Markov model (HMM), hybrid deep neural network (HMM-DNN) and attention-based encoder-decoder (AED) ASR recipes. We set benchmarks on the official test sets, as well as multiple other recently used test sets. Both temporal corpus subsets are already large, and we observe that beyond their scale, ASR performance on the official test sets plateaus, whereas other domains benefit from added data. The HMM-DNN and AED approaches are compared in a carefully matched equal data setting, with the HMM-DNN system consistently performing better. Finally, the variation of the ASR accuracy is compared between the speaker categories available in the parliament metadata to detect potential biases based on factors such as gender, age, and education.
\end{abstract}

\newpage

\section{Introduction}
\label{intro}

Automatic Speech Recognition (ASR) training data, transcribed speech, is expensive to create. There are some publicly funded large-scale efforts to specifically create this type of data, to facilitate language technologies in the chosen language. Commercial agents create private datasets, driven by business ventures and internal research projects. Lastly, there is data, both public and private, which has not been explicitly created for ASR research, but which can be leveraged for that purpose. A common source in many languages have been certain public forums such as parliaments, which often produce and publish transcripts of their sessions.

This work presents, to the best of our knowledge, the largest public monolingual corpus of parliament session data purposed for ASR, and the largest transcribed public Finnish ASR corpus, at a little over 3000 hours altogether. We present benchmark results and provide recipe starting points for the data. Additionally, we explore how models trained on parliament session speech generalise to other domains and modes of speech. The extent of the data lends itself to comparisons between Hidden Markov Model (HMM) systems, which are known to already work well at smaller scales and end-to-end models, which are thought to become competitive at larger scales.

New Finnish parliament session data becomes available constantly, and our data collection and processing pipeline can be run intermittently, producing new ever larger versions of this corpus. This prompts the question: how much improvement in ASR performance can we expect from more data? Furthermore, how does the data distribution shift in time - do models trained on data from previous electoral cycles perform well on new data, which has new voices and new topics such as the COVID-19 pandemic?

The Finnish Parliament provides a rich set of metadata about the speakers and recordings. These allow us to statistically analyse the dataset and our ASR results on it from multiple viewpoints. Furthermore, the sessions are also captured on video, and thus in this work we wish to lay the ground work for future research in multimodal, audiovisual tasks.

In summary, the contributions of this work are:
\begin{itemize}
    \item Public release of the largest transcribed Finnish ASR dataset (3000 hours of manually transcribed and automatically segmented and aligned speech), with two temporally distinct subsets, and two test sets
    \item Development of new pipeline for retrieving and processing Finnish parliament recordings and transcripts
    \item Benchmark results, recipe starting points, generalisation study, and an equal data comparison on hidden Markov model systems and end-to-end Attention-based encoder-decoder models
    \item Analysis of the results as well as the data based on multiple metadata factors
\end{itemize}

\section{Related work}
\label{sec:background}

\begin{table}[t!]
\centering
\caption{Properties of datasets named in section~\ref{sec:background}. Lahjoita Puhetta and VoxPopuli have both transcribed and untranscribed subsets. The speaker count for untranscribed VoxPopuli was not defined in the paper~\citep{wang-2021-voxpopuli}. For Speecon and SpeechDat we report the size and speaker count of the Finnish subsets.}
\label{tab:datasets}
\begin{tabular}{lrrll}
\hline\noalign{\smallskip}
\textbf{Data set}      & \textbf{Size} & \textbf{Speakers} & \textbf{Languages} & \textbf{Domain} \\
\noalign{\smallskip}\hline\noalign{\smallskip}
\multicolumn{5}{c}{\textit{Finnish datasets}} \\
\noalign{\smallskip}\hline\noalign{\smallskip}
DSPCon                           & 10 h             & 242    & fi    & conversational \\ 
FinDialogue                      & 10 h             & 22     & fi    & conversational \\ 
Finnish parliament               & 3 087 h          & 449    & fi    & political \\
Lahjoita puhetta                 & 1 601 h          & 17 821 & fi    & spontaneous \\ 
Lahjoita puhetta (untranscribed) & 1 597 h          & 18 825 & fi    & spontaneous \\ 
Speecon                          & 204 h            & 550    & 20    & read, spontaneous \\
SpeechDat                        & 236 h            & 4 000  & 14    & read \\
\noalign{\smallskip}\hline\noalign{\smallskip}
\multicolumn{5}{c}{\textit{Parliament datasets}} \\
\noalign{\smallskip}\hline\noalign{\smallskip}
Bern parliament                  & 293 h            & 224    & de     & political \\
Bulgarian parliament             & 249 h            & 572    & bg     & political \\
Czech parliament                 & 444 h            & 212    & cs     & political \\
Danish parliament                & 1 857 h          & 434    & da     & political \\
Finnish parliament               & 3 087 h          & 449    & fi     & political \\
Icelandic parliament             & 542 h            & 197    & is     & political \\
Valais parliament                & 40 h             & 204    & de, fr & political \\
VoxPopuli                        & 1 791 h          & 4 295  & 16     & political \\
VoxPopuli (untranscribed)        & 384 000 h        & -      & 23     & political \\
\noalign{\smallskip}\hline
\end{tabular}
\end{table}

\cite{mansikkaniemi2017automatic} created and studied the first version of Finnish parliament ASR corpus.
They extracted 1559 hours of ASR data from over 2000 hours of video recordings.
This data was never publicly released and the retrieval pipeline used web scraping and obsolete interfaces. The data preparation pipeline was based on AaltoASR\footnote{\url{https://github.com/aalto-speech/AaltoASR}}. This preliminary work forms one training subset of our data, as well as contributing the temporally earlier test set, and the development set. We develop a new retrieval pipeline for the new official open data interface and base our processing pipeline on Kaldi~\citep{povey2011kaldi}. We retrieve all data available on the new interface, which forms the second, more recent training subset in our corpus. In addition, we pool all data together to form a combined training subset of 3087 hours. We manually correct and curate a new test set, which covers a later period of time compared to the first test set. We report new benchmark HMM-DNN results, and additionally, we report AED benchmark results, and publish HMM-DNN and AED recipes. We analyse the data and model errors and biases statistically.

In Table \ref{tab:datasets}, we can see a list of relevant Finnish corpora. Before the Finnish parliament ASR corpus, early large-scale efforts to collect Finnish speech data were part of larger international projects to build multilingual speech databases - SpeechDat for speech-driven teleservices~\citep{hoge1999speechdat} and Speecon for speech-driven interfaces in consumer devices~\citep{iskra-2002-speecon}. Both databases are around 200 hours in size and contain mainly read speech like isolated digits and words, numbers, spellings, dates, commands, and sentences. These databases are publicly available for academic and commercial use, but only for fees ranging from 30\,000 to 75\,000 euros.

DSPCon \citep{enarvi2018}, FinDialogue~\citep{lennes2009finintas}, and Lahjoita puhetta~\citep{moisio2022lahjoita} corpora represent more spontaneous and conversational forms of Finnish speech. DSPCon contains short conversations between students. Similarly, the FinDialogue corpus is a collection of 10 spontaneous long dialogues between friends. Lahjoita Puhetta is a new considerably larger corpus of spoken and colloquial Finnish made of speech donations collected from the general public. The corpus has both transcribed and untranscribed subsets and covers a large variety of speakers and speaking styles. There are also more databases of spontaneous Finnish, such as Prosovar, but they are currently not publicly available~\citep{lindenCLARINbook}.

Besides the Finnish parliament data, parliament meeting records and transcripts have provided a valuable source of ASR data for many other languages as well. We list examples given here for comparison in Table \ref{tab:datasets}. One of the earliest examples is the MediaParl corpus for French and German spoken in the Swiss Valais Parliament by \cite{imseng2012mediaparl}. In recent years, public corpora based on parliament records has also been created for Icelandic~\citep{helgadottir2017building}, Bulgarian~\citep{geneva2019building}, Danish~\citep{kirkedal2020ftspeech}, Czech~\citep{kratochvil-2020-large}, and Swiss German~\citep{pluss2020swiss}. Various event recordings from the European parliament have also been used to create the large multi-lingual VoxPopuli ASR dataset which contains both transcribed and untranscribed speech data~\citep{wang-2021-voxpopuli}.

\section{Data preparation}
\label{sec:pipeline}

The speech in Finnish parliament sessions is a mix of planned speech, like opening statements and interpellations, and more spontaneous speech such as debate. The 200 members of parliament, of which 94 are women in 2022, are elected every four years from 13 districts around Finland. Parliament sessions are held four times a week and each working year is split to two terms - spring term from February to June and autumn term from September to December. Yearly roughly 500 hours of new video recordings become available.

The data preparation pipeline for the Finnish parliament speech has been completely redone since the previous iteration described in \cite{mansikkaniemi2017automatic}. The reasons for this are two-fold. First, since the previous iteration, the Finnish parliament has made changes to their open data interfaces. The first version spanned years 2008-2016 and transcripts were crawled directly from HTML pages. Starting from 2015, the plenary transcripts have been available in a rich XML format with more metadata such as language labels and member of parliament (MP) id. To use all that metadata, the new dataset spans years 2015-2020. The second reason is to move to the Kaldi toolkit~\citep{povey2011kaldi} which can be used to implement state-of-the-art models, and has a well-tested set of segmentation tools\footnote{\url{https://github.com/kaldi-asr/kaldi/blob/master/egs/wsj/s5/steps/cleanup}} developed by \citet{manohar2017jhuMGB}. The full pipeline is made available on Github.\footnote{\url{https://github.com/aalto-speech/fi-parliament-tools}}

\subsection{Challenges of the data}

The primary challenge of exploiting the plenary sessions as speech recognition data is the length of the plenary recordings. They vary from 15 minutes to 18 hours in length. However, data samples used to train ASR are generally less than 30 seconds long~\citep{chiu2019long-form}. Computational challenges have limited the length for statistical models in the past~\citep{meyer2006boostingHMM} and continue to do so in the contemporary neural network models~\citep{chiu2019long-form}. Therefore, we need to segment the sessions into smaller pieces more suitable for ASR training.

A second challenge is formed by mismatches between audio and transcripts because the transcripts are edited for clarity and readability. Furthermore, hesitations, repetitions, and colloquial pronunciations are omitted. \cite{voutilainen2017fiparl} states that in the Finnish parliament transcripts self-corrections, slips of the tongue and selected particles are edited for readability. Additionally, morphological and syntactic features of spoken and spontaneous language are replaced with equivalent written language. For the morphological case for instance, Voutilainen writes that \textit{me mennään} is changed to \textit{me menemme} ('we go').

There are further complications as well. For instance, some speech is left untranscribed or the speaker is not clearly marked. In our dataset, we wanted to include only speech where the speaker is known so we needed to be able to skip other speech. The recording may also contain long silent sequences where the camera films the room but microphones are muted if the meeting has started late or there are breaks. Finally, transcripts are not always ordered chronologically causing mismatch with the audio. Transcripts always follow the agenda of the day, but in long sessions the chairman may choose to first discuss topics which incur less debate.

\subsection{Pipeline steps}

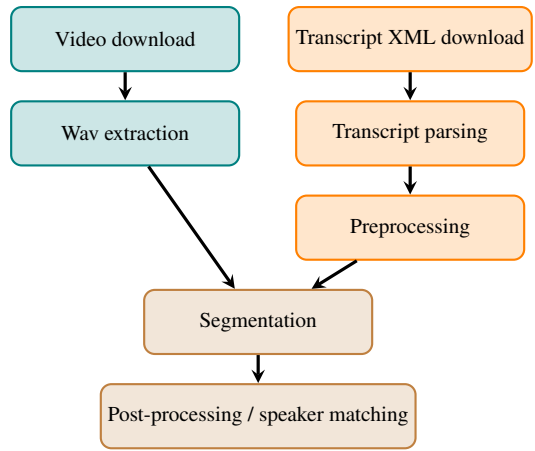
\begin{figure}[t]
    \centering
    \begin{tikzpicture}[
    node distance=1.25cm,
    video/.style = {rectangle, rounded corners, thick, minimum width=3cm, minimum height=0.85cm,text centered, draw=teal, fill=teal!20},
    txt/.style = {rectangle, rounded corners, thick, minimum width=3cm, minimum height=0.85cm,text centered, draw=orange, fill=orange!20},
    both/.style = {rectangle, rounded corners, thick, minimum width=3cm, minimum height=0.85cm,text centered, draw=brown, fill=brown!20},
    arrow/.style = {very thick, ->, >=stealth}
    ]
    \node (video_dl) [video] {Video download};
    \node (wav_extract) [video, below of=video_dl] {Wav extraction};
    \node (xml_dl) [txt, right of=video_dl, xshift=2.5cm] {Transcript XML download};
    \node (parsing) [txt, below of=xml_dl] {Transcript parsing};
    \node (preprocessing) [txt, below of=parsing] {Preprocessing};
    \node (segmentation) [both, below of=preprocessing, xshift=-2cm] {Segmentation};
    \node (postp) [both, below of=segmentation] {Post-processing / speaker matching};
    \draw [arrow] (video_dl) -- (wav_extract);
    \draw [arrow] (xml_dl) -- (parsing);
    \draw [arrow] (parsing) -- (preprocessing);
    \draw [arrow] (wav_extract) -- (segmentation);
    \draw [arrow] (preprocessing) -- (segmentation);
    \draw [arrow] (segmentation) -- (postp);
    \end{tikzpicture}
    \caption{A flowchart showing the steps in the data preparation pipeline.}
    \label{fig:pipeline}
\end{figure}

The steps of data preparation pipeline are visualised in Figure~\ref{fig:pipeline}. It begins with the download of the plenary session videos from the video service provider and corresponding XML transcripts from the parliament open data API. After the downloads, a standard 16 kHz single-channel audio wave file is extracted from the video using the ffmpeg tool\footnote{\url{https://ffmpeg.org}}. From the XMLs, we parse speech transcripts, speaker name, member of parliament id, language (Finland is bilingual), political party, title (e.g. chairman, prime minister), and approximate timestamps. Not all of the speeches have the full metadata, but the speaker name and title are always there. Parsed data is saved in a JSON file for human-readability and interoperability.

The second step is to preprocess the speech transcripts for the Kaldi segmentation script. For each plenary session, all speech transcripts are preprocessed into one long line of text. The preprocessing maps all latin characters outside the Finnish alphabet to their closest equivalent in the Finnish alphabet, for example ø to ö. It also removes transcribed exclamations from other MPs and punctuation, expands digits and abbreviations, and lowercases all text. Since punctuation is removed, samples generated by the segmentation script can start and end in the middle of a sentence. This is in contrast to \citep{mansikkaniemi2017automatic}, where segmentation was based on sentence boundaries. During the preprocessing, we also fill in the language labels and speaker IDs in the JSONs if they are missing. Speech is labelled as Finnish or Swedish or both using predictions from FastText's language identification model~\citep{joulin2017bag}. We differentiate the predicted labels from gold standard labels in the JSONs by adding \textit{.p} after the label. Speaker ids are looked up using the speaker name from a speaker metadata table collected from the parliament API.

In the third step, audio files and preprocessed texts are segmented with the Kaldi segmentation script. Since speaker turn changes were lost each session was preprocessed to single long line of text, the output of the segmentation script needs to be matched to the JSON transcript for speaker retrieval. For each speech in a plenary transcript JSON, we search for it in the segmented time-marked conversation file (CTM) generated by the segmentation script. If the speech is found, we mark the speaker and the statement language in the CTM. After matching is done, we keep only segments that have one speaker and the language is Finnish.

\subsection{Published corpus}
\label{sec:corpus}

With the new pipeline, we processed plenary meetings from 2015 until end of June 2020, a total of 743 sessions. All in all, the raw audio from these sessions was 2448 hours long, of which 1783 hours, or approximately $73\%$, ended up in the final \textit{Train20} training dataset. Out of the raw data, 22 \% was lost in the segmentation process, due to three causes: 1) the audio was silent e.g. microphone was muted, 2) the speech was not transcribed, or 3) the ASR model could not recognise the speech accurately enough for alignment with transcript. 5 \% of the raw data was lost in the post-processing step, when speaker identities were recovered. We discarded samples that had more than one speaker as well as samples that contained speech in Swedish.

The size of the previous, \textit{Train16} \citep{mansikkaniemi2017automatic}, training dataset is 1559 hours and it covers sessions from autumn 2008 to summer 2016. To combine the two datasets, we needed to remove overlapping samples. We decided to drop samples spanning years 2015 and 2016 from the old parliament set, so that the \textit{Combined} dataset would cover years 2008-2020. From the new, \textit{Train20} dataset, we removed any samples that overlapped with development or test sets defined in \cite{mansikkaniemi2017automatic}. When combined, we got a training corpus with 3087 hours of data and 449 speakers. The full data is made available in Kielipankki\footnote{\url{http://urn.fi/urn:nbn:fi:lb-2021081105}}.

For model development and testing, we use the development and test sets from \cite{mansikkaniemi2017automatic}. In addition, we create a new test set to evaluate domain shift in the parliament speeches. The new, \textit{Test20} set, is sampled from the segmented plenary meetings of autumn 2020. We listen and correct all samples in this set by hand. Details of all the parliament data subsets are listed in Table~\ref{tab:parl-data}.

\begin{table}[t]
\centering
\caption{Size, speaker count, and abbreviation of the different Finnish parliament ASR corpus subsets.}
\label{tab:parl-data}
\begin{tabular}{llrr}
\hline\noalign{\smallskip}
\textbf{Subset name}     & \textbf{Abbreviation} & \textbf{Size} & \textbf{Speakers} \\
\noalign{\smallskip}\hline\noalign{\smallskip}
Train 2008-2016          & Train16               & 1559 h        & 357               \\
Train 2015-2020          & Train20               & 1783 h        & 302               \\
Combined train 2008-2020 & Comb                  & 3087 h        & 449               \\
Development 2008-2016    & Dev16                 & 5 h           & 19                \\
Test 2008-2016           & Test16                & 5 h           & 21                \\
Test 2020                & Test20                & 5 h           & 28                \\
\noalign{\smallskip}\hline
\end{tabular}
\end{table}

\begin{figure}[hp!]
    \centering
    \begin{subfigure}[b]{0.6\textwidth}
        \includegraphics[width=\textwidth]{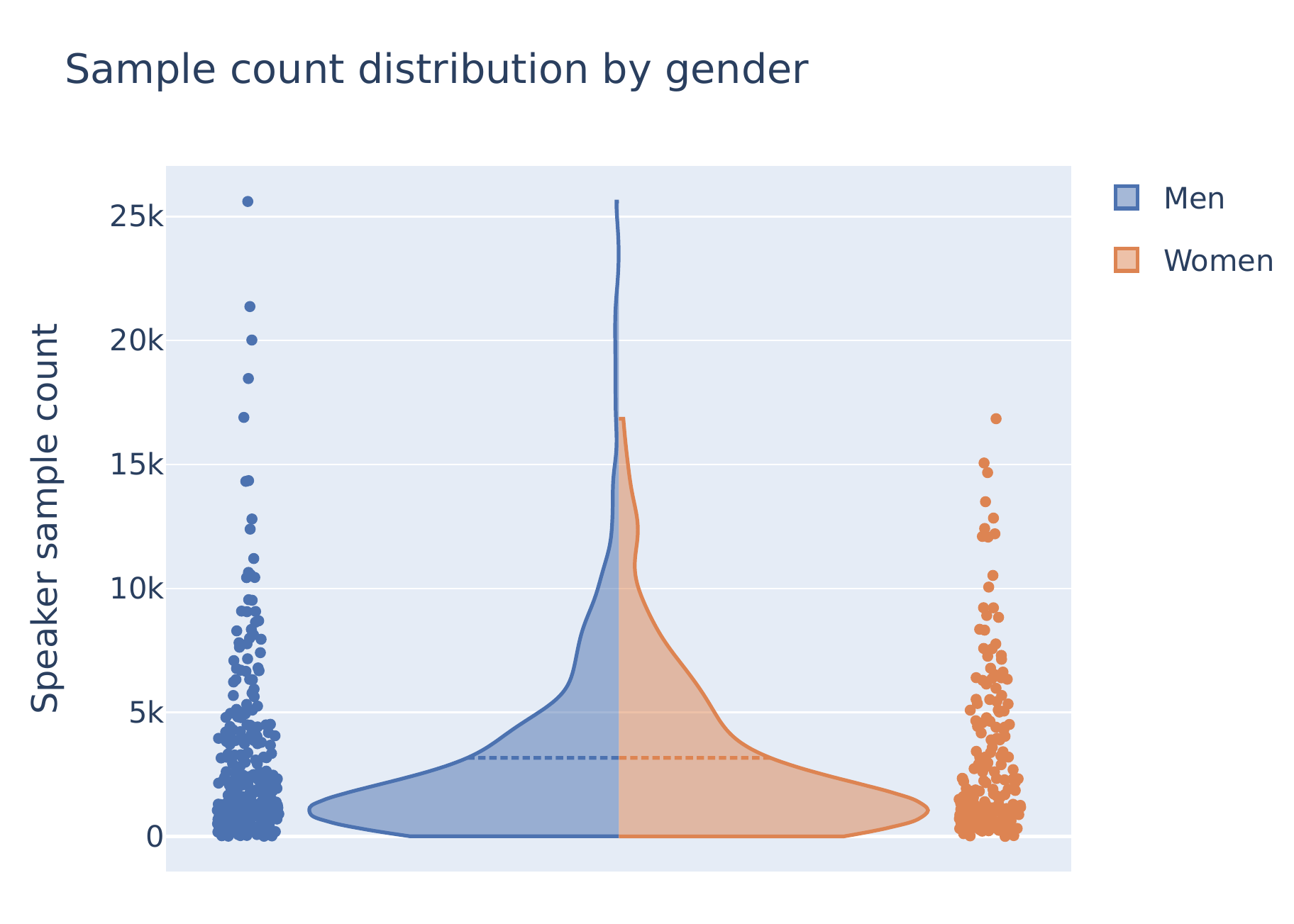}
        \caption{Men have more speakers with very high sample count.}
        \label{fig:data_stats_speaker_count}
    \end{subfigure}
    \hfill
    \begin{subfigure}[b]{0.99\textwidth}
        \includegraphics[width=\textwidth]{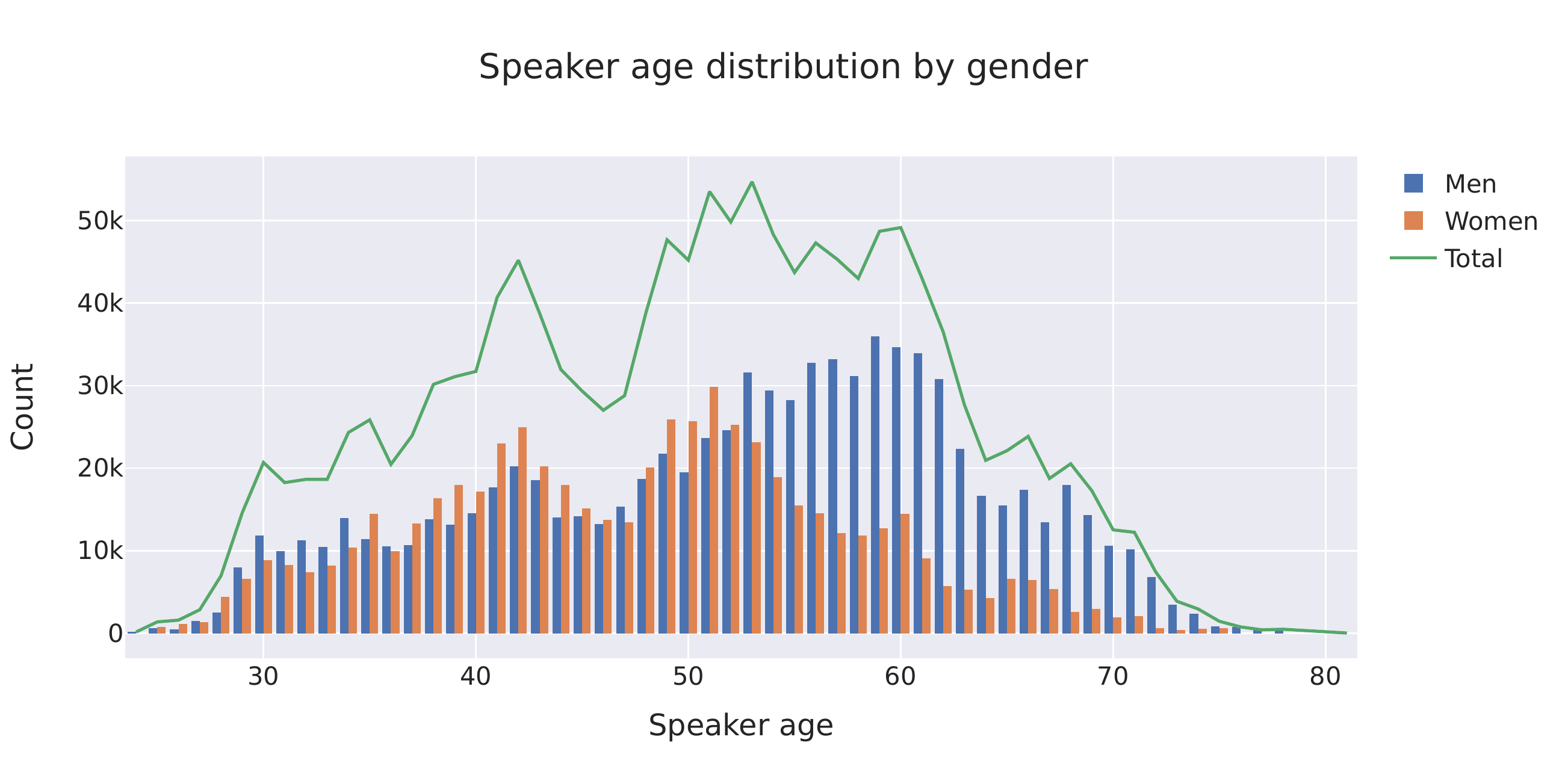}
        \caption{Women form a smaller share of samples in over 50-year-old speakers.}
        \label{fig:data_stats_age}
    \end{subfigure}
    \hfill
    \begin{subfigure}[b]{0.99\textwidth}
        \includegraphics[width=\textwidth]{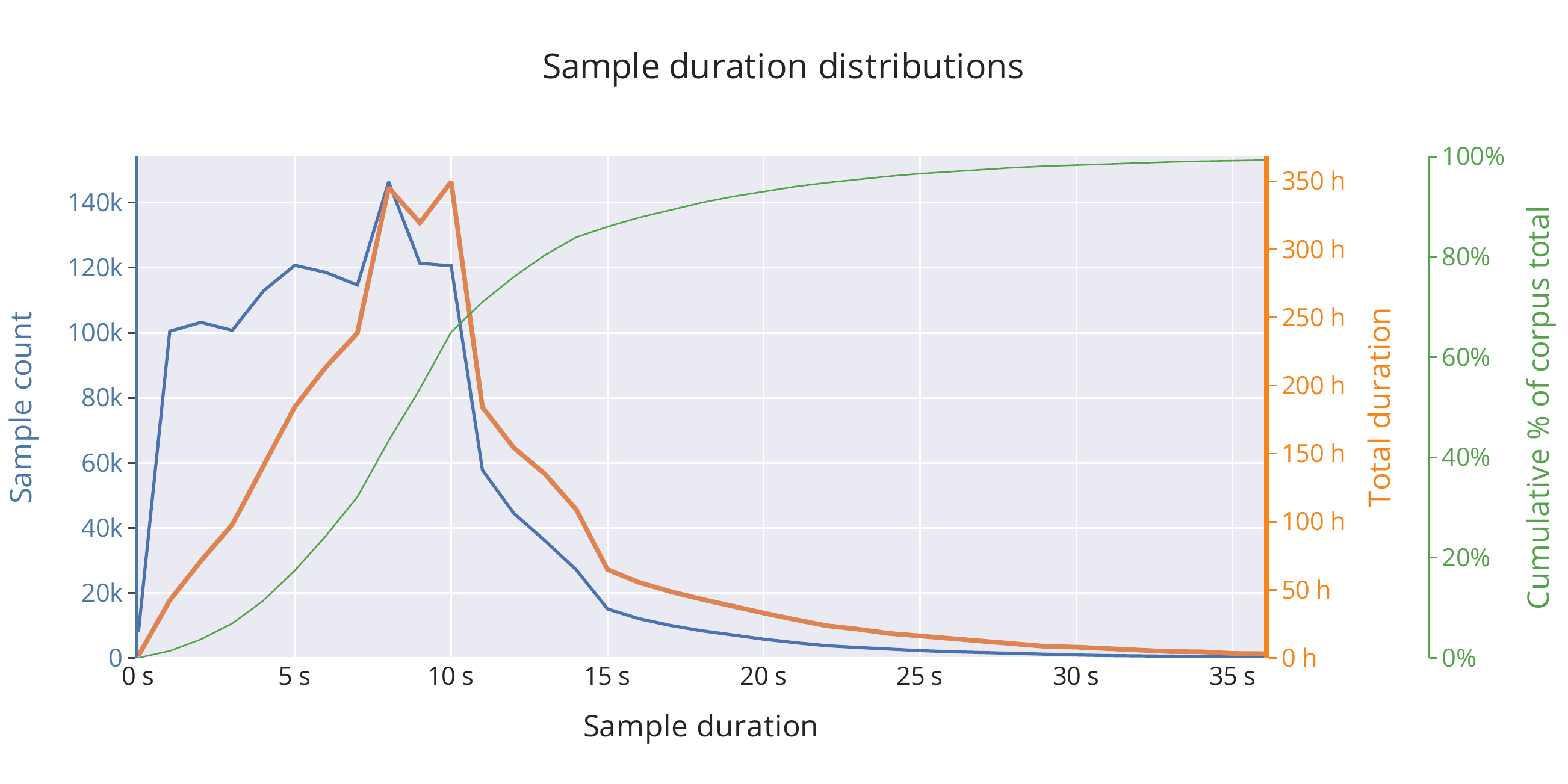}
        \caption{Over 86~\% of the samples are 15 seconds or less in length.}
        \label{fig:data_stats_dur}
    \end{subfigure}
    \caption{Some characteristics of the \textit{Combined} 2008-2020 training set.}
    \label{fig:data_stats}
\end{figure}

The gender, age and duration statistics of the \textit{Combined} dataset are visualised in Figure~\ref{fig:data_stats}. First, Figure~\ref{fig:data_stats_speaker_count} shows how the samples are distributed among speakers and gender. Majority of the speakers have less than 5 000 samples each, but there are a few outliers especially among male speakers. Overall, women account for 40~\% of both the speakers and the speech audio. 

Age-wise, the samples are distributed evenly between genders for speakers below 53 years but men dominate among older speakers, as is shown in Figure~\ref{fig:data_stats_age}. What is also notable is that, in most ASR datasets, each speaker gives their sample at a certain age. But in this dataset samples from a single speaker come from a range of years, since MPs usually serve a full electoral term or more. 

Different distributions of sample duration are shown in Figure~\ref{fig:data_stats_dur}. Over 85 \% of the samples are 15 seconds or shorter in length. This partially due to the default parameters of the Kaldi segmentation script, which sets the maximum sample length to 15 seconds. Thus, the newer \textit{Train20} set contains only samples up to 15 seconds, while the \textit{Train16} set contains longer samples.

\section{Models}
\label{sec:models}

Besides simply demonstrating and evaluating the basic uses of this new data resource, our speech recognition experiments have three main goals. Firstly, these experiments provide benchmarks and recipe starting points for new research. Secondly, we demonstrate how the new \textit{Train20} resource complements the \textit{Train16} set. Finally, we explore how models trained on this data generalise to other existing test sets.

We start our experiments by optimising Gaussian Mixture Model (GMM) acoustic models. GMMs are no longer an area of active research, but they are typically needed in Hidden Markov Model (HMM) -based speech recognition to generate and refine alignments, and to cluster the context-dependent phone HMM states. We improve the multi-stage GMM recipe over 27 runs, so that future work on this data can build on a ready, optimised GMM recipe without needing to redo this additional effort. We also provide a benchmark Deep Neural Network (DNN) acoustic model and Attention-based Encoder-Decoder (AED) system. Recipes for the experiments are made available in Github\footnote{\url{https://github.com/aalto-speech/fin-parl-models}}.

\subsection{HMM-GMM}
\label{sec:hmm-gmm}

The outline of our GMM recipe is the Kaldi~toolkit~\citep{povey2011kaldi} standard. All the GMMs use a tri-state HMM-topology with probability density functions (PDF) tied through a phonetic decision tree. As inputs we use 13 Mel-frequency cepstral coefficients (MFCCs). The recipe begins with a monophone model that is trained with the shortest 2000 samples. The monophone alignments are then used to train the first $\Delta$+$\Delta\Delta$ triphone model with 100\,000 randomly selected utterances. The second triphone model splices together seven feature vectors through a Linear Discriminant Analysis (LDA) transform as its input, and adjusts the model through a Maximum Likelihood Linear Transform (MLLT). It is trained with 250\,000 randomly selected utterances. The third and fourth triphone models add speaker adaptive training (SAT) on top of LDA and MLLT. The former is trained with the same 250\,000 samples while the latter trained on the whole data set.

In addition to the basic Kaldi recipe, we tried other configurations as well, like leaving out one or two of triphone training steps or training from the beginning with full data set. However, the results did not improve.

The fifth model is the final HMM-GMM that is used to generate alignments for the neural network training in the next stage. We tuned the number of PDFs (leaves) and the number of Gaussians and updated the model until convergence at 70 iterations. The best result for the \textit{Train20} set used 12\,304 leaves in the phonetic decision tree and 501\,717 Gaussians - the recipe hyperparameters being 14\,000 maximum number of leaves and 500\,000 target number of Gaussians.

\subsection{Time delay neural networks (TDNN)}
\label{sec:tdnn}

Our DNN acoustic model benchmark is based on the Kaldi Librispeech~recipe\footnote{\url{https://github.com/kaldi-asr/kaldi/blob/66f5434/egs/librispeech/s5/local/chain/tuning/run_tdnn_1d.sh}}. To make the recipe simpler and easier to compare against, we did not use i-vectors nor speed perturbation.
The inputs for this model are 80 Mel-scale filterbank features. The model has 16 factorised~Time~Delay~Neural~Network~(TDNN-F) layers~\citep{povey18_interspeech} as well as an initial LDA splicing layer and a final feed-forward layer. The model has 18\,800\,000 parameters and was trained for 8 epochs, as described by \citet{Povey2016lfmmi}, with the Lattice-free Maximum Mutual Information criterion and a regularising Cross-Entropy criterion.

We train a smaller model of 11 TDNN-F layers (7.5M parameters) and a larger model of 20 TDNN-F layers (26.7M parameters), but in our preliminary experiments, they perform the same or worse. In addition to plain TDNN models, we also test an architecture with three TDNN-F layers and three Bidirectional Long Short-Term Memory (BLSTM) layers with a total of 46.1 million parameters.

\subsection{Language models and lexicon}
\label{sec:lm}

Because the agglutinative nature of Finnish would require a very large word lexicon,
we opt for a subword-based solution. We develop several n-gram language models (LM) with byte-pair-encoding (BPE) subwords~\citep{sennrich2016bpe} on both in-domain and out-of-domain data using the VariKN~\citep{siivola2007varikn} and SentencePiece~\citep{kudo-richardson-2018-sentencepiece} tools. Neural network language models are left for future work as they are out of scope in this work. All LMs presented here were trained up to 10-grams with a scaling factor of $0.0001$, unless otherwise stated. We use a grapheme-based lexicon in this work because Finnish has near phonemic orthography.

For each acoustic model, we train a matching language model from the training transcripts with 1 750 BPE units. We also train a larger, 19\,000 BPE unit, in-domain language model from the 20M token parliament meeting transcript corpus collected in~\citep{mansikkaniemi2017automatic}. This corpus contains full sentences parsed from the meeting transcripts. It is based on the same plenary meetings data, 2008-2016, as the ASR corpus, so we make a new extended version of it covering meeting transcripts from 2008 to mid-2020. Using this extended 30M token corpus, we train a second in-domain language model with 19\,000 BPE units.

In addition to in-domain models, we use two out-of-domain language models to evaluate the new acoustic models on out-of-domain test data. We train a general-domain LM with 19 000 BPE units with the Kielipankki corpus\footnote{\url{http://urn.fi/urn:nbn:fi:lb-201403268}}. It is a collection of Finnish texts from 1990s newspapers, journals and books. As this is a large 144M token dataset, we train both a smaller 5-gram LM for the first recognition pass to create lattices and a 10-gram model for rescoring the lattices. The second model is a word-based, Kneser-Ney smoothed 4-gram for conversational Finnish developed by \cite{enarvi2017convfinnish}.

\subsection{Attention-based Encoder-Decoder models}
\label{sec:e2e}

Various end-to-end models have recently become mainstream approaches in ASR. We train end-to-end Attention-based Encoder-Decoder (AED) models~\citep{bahdanau2016attention,chan2016las} to explore this direction. We implement the models in the SpeechBrain toolkit~\citep{speechbrain}. The encoder is a stack of convolutional, Long Short Term Memory (LSTM), and feed-forward layers, with four-fold frame subsampling in the convolutional layers. The attention mechanism is a content-and-location aware variant, and the decoder is a stack of Gated Recurrent Units (GRU). The model takes 40-element Mel-Filterbank log-energy vectors as input and computes a distribution over a vocabulary of 1750 SentencePiece BPE-units. The model has 27.7M parameters. It is trained with dynamic batching, targeting 40 seconds of audio per batch, for 100 nominal epochs of 10\,000 updates, with early stopping. The first 15 nominal epochs use an auxiliary Connectionist Temporal Classification (CTC) loss on the encoder to aid initial convergence~\citep{jointattentionctc}. In preliminary experiments, we tried to improve on this benchmark model by building various larger models, but none yielded better results.

We take care to compare the AED models against HMM-systems in an \textit{equal data setting}~\citep{rouhe2021equal}: both paradigms use just transcribed speech as the training data. Since typically HMM-systems leverage additional text data and expert lexica, comparing them with end-to-end models which are only trained on transcribed speech confounds differences in models and learning with differences in the training data. Using just the transcripts for language modeling still follows standard practices, and in Finnish, grapheme lexica are also standard practice, due to the transparent orthography. Further care is taken to balance the comparison by using a matching vocabulary with both paradigms and not using augmentation nor i-vectors with either paradigm.

AED models are known to struggle with long-form speech~\cite{chiu2019long-form}. When applying the AED models on our out-of-domain Lahjoita Puhetta test data, which has long-form recordings, some utterances produce pathological repetitive output, similar to reports by \citet{keung2020attentional}. To eliminate that we apply a simple post-processing filter where we allow repetitions to produce in total a maximum of five tokens.

\section{Results}
\label{sec:results}

In this paper we present only the main ASR word error rate (WER) results, more can be found in Github\footnote{\url{https://github.com/aalto-speech/fin-parl-models}}. First, we compare acoustic models (AM) trained with different subsets of training data to study how much the increased training data improves the performances. We continue with a similar comparison of language models (LM) trained with different amounts of in-domain text data. Then we evaluate the best TDNN model on various Finnish ASR benchmark test sets. Finally, we evaluate AED models on the same benchmarks and compare the results to a HMM-system trained in an equal data setting, which here simply means limiting the HMM-system to the transcript LM.

\subsection{Model development}
\label{sec:results-parl}

We started our model development efforts from the HMM-GMM optimisation on the \textit{Train20} set. We use the development set \textit{Dev16} for tuning all the model hyperparameters. The results for the best HMM-GMM are in Table~\ref{tab:hmm-and-tdnn-data-comparison}. 

When we compare \textit{Train20} HMM-GMMs to the \textit{Train16} and \textit{Combined} sets on our development set, we see that \textit{Combined} set gives a small improvement as expected. However, with DNN models the \textit{Combined} set TDNN-medium performs worse than the same model using \textit{Train20}. Even with the over twice as big TDNN-BLSTM model (see AM parameter counts in Table~\ref{tab:lm-comparison}), where we assumed more data would help, the improvement is statistically insignificant. 

\begin{table}[t!]
\centering
\caption{\textit{Dev16} WERs of acoustic models trained with different subsets of the training data and evaluated with the 20M token in-domain LM. Results in WER~[\%].}
\label{tab:hmm-and-tdnn-data-comparison}
\begin{tabular}{lrrrr}
\hline\noalign{\smallskip}
\textbf{Acoustic model} & \textbf{Data size (samples)}  & \textbf{Train20} & \textbf{Train16} & \textbf{Comb} \\
\noalign{\smallskip}\hline\noalign{\smallskip}
Monophone GMM           & 2k   & 56.24         & 69.87         & 61.29 \\
Delta+delta-delta GMM   & 100k & 21.56         & 21.43         & 21.34 \\
LDA+MLLT GMM            & 250k & 17.83         & 17.72         & 17.63 \\
LDA+MLLT+SAT GMM        & 250k & 16.70         & 16.77         & 16.41 \\
LDA+MLLT+SAT GMM        & all  & 14.34         & 14.42         & 14.09 \\
TDNN-medium             & all  & \textbf{9.98} & 10.34         & 10.28 \\
TDNN-BLSTM              & all  & 10.66         & 11.06         & 10.54 \\
\noalign{\smallskip}\hline\noalign{\smallskip}
\multicolumn{4}{l}{\textit{After cleanup}} \\
\noalign{\smallskip}\hline\noalign{\smallskip}
LDA+MLLT+SAT GMM        & all  & 14.31         & 14.22         & 14.01 \\
TDNN-medium             & all  & 9.37          & \textbf{8.49} & 8.69  \\
\noalign{\smallskip}\hline
\end{tabular}
\end{table}

In the work of \cite{mansikkaniemi2017automatic}, experiments are performed with data that is cleaned using the Kaldi cleanup tools. To gauge the effect the cleanup has on the training subsets, we run the cleanup script for each with the final, full data HMM-GMM model as is the Kaldi standard. With the cleaned data, we train another LDA+MLLT+SAT HMM-GMM and TDNN-medium model. For the HMM-GMM models the cleanup brings negligible improvements but for the TDNN-medium models improvement is clear. The biggest improvement is seen in the \textit{Train16} set which we believe is due to the segmentation process being different from the Kaldi standard. The smaller improvement for the \textit{Train20} set implies the cleanup is less important for this set. Furthermore, the \textit{Combined} set demonstrates the same saturation effect as without cleanup.

For the \textit{Train20} set, we train a smaller and a larger TDNN model to see how model size influences the performance on this data. The results are shown in Table~\ref{tab:lm-comparison}. It appears that the benchmark TDNN-medium is already a good fit for the \textit{Train20} set. Additionally, in the same table, we compare the three in-domain language models. Transcripts make a decent language model, but adding available in-domain data is even better. 

On the Parliament tasks, a saturation effect is observed. The acoustic model does not improve after increasing the training data from the \textit{Train20} subset to the full \textit{Combined} set. Similarly, the language model does not improve after increasing the training data from the 20M subset to the full 30M data.

\begin{table}[!t]
    \caption{In-domain language model comparisons for uncleaned \textit{Train20} acoustic models on \textit{Dev16} development set. Acoustic model parameters are in millions and results in WER~[\%].}
    \label{tab:lm-comparison}
    \centering
    \begin{tabular}{lrrrr}
        \hline\noalign{\smallskip}
        \textbf{Acoustic model} & \textbf{Parameters} & \textbf{Parl-20M} & \textbf{Parl-30M} & \textbf{Transcript} \\
        \noalign{\smallskip}\hline\noalign{\smallskip}
        Final GMM               & 0.5M                & 14.34             & 14.38             & 21.12                \\
        TDNN-small              & 7.5M                & 10.24             & 10.22             & 15.19                \\
        TDNN-medium             & 18.8M               & 9.98              & \textbf{10.02}    & \textbf{14.19}       \\
        TDNN-large              & 26.7M               & \textbf{9.97}     & 10.06             & 14.34                \\
        TDNN-BLSTM              & 46.1M               & 10.66             & 10.58             & 14.34                \\
        \noalign{\smallskip}\hline
    \end{tabular}
\end{table}

\subsection{Test set results}
\label{sec:testset-results}

Next, we focus on the TDNN-medium model and evaluate it on five test sets using the three uncleaned acoustic model training sets and five language models. Results are displayed in Table~\ref{tab:testset-lm-comparison}. Two of the test sets, \textit{Test16} and \textit{Test20}, are in-domain while remaining three are out of domain. Lahjoita puhetta test set contains spontaneous and colloquial speech, Speecon consists of mainly read speech in various conditions, and YLE test set is made of news and broadcast material. Language models were detailed in section~\ref{sec:lm}.

\begin{table}[t]
\centering
\caption{Comparison of test set results for different language models and uncleaned acoustic model training sets on the TDNN-medium acoustic model. Results in WER~[\%].}
\label{tab:testset-lm-comparison}
\begin{tabular}{lrrrrr}
\hline\noalign{\smallskip}
\textbf{AM training data}     & \textbf{Parl-20M} & \textbf{Parl-30M} & \textbf{Transcript} & \textbf{General} & \textbf{Conversational} \\
\noalign{\smallskip}\hline\noalign{\smallskip}
\multicolumn{6}{c}{\textit{Test16 set}} \\
\noalign{\smallskip}\hline\noalign{\smallskip}
Train20            & \textbf{6.97} & 7.05           & 10.52         & 10.92          & 17.83          \\
Train16            & 7.83          & \textbf{7.77}  & 11.17         & 11.77          & 18.89          \\
Combined           & \textbf{7.14} & 7.15           & 9.83          & 10.60          & 17.59          \\
\noalign{\smallskip}\hline\noalign{\smallskip}
\multicolumn{6}{c}{\textit{Test20 set}} \\
\noalign{\smallskip}\hline\noalign{\smallskip}
Train20            & 9.83          & 9.34           & \textbf{8.84} & 9.59           & 17.43          \\
Train16            & 14.97         & \textbf{13.17} & 13.90         & 13.89          & 21.29          \\
Combined           & 10.67         & 9.73           & \textbf{8.76} & 10.12          & 17.57          \\
\noalign{\smallskip}\hline\noalign{\smallskip}
\multicolumn{6}{c}{\textit{Lahjoita puhetta test set}} \\
\noalign{\smallskip}\hline\noalign{\smallskip}
Train20            & 66.59         & 66.20          & 66.90         & 64.85          & \textbf{60.05} \\
Train16            & 68.85         & 67.26          & 68.41         & 64.73          & \textbf{58.82} \\
Combined           & 65.48         & 65.16          & 64.99         & 62.79          & \textbf{57.42} \\
\noalign{\smallskip}\hline\noalign{\smallskip}
\multicolumn{6}{c}{\textit{Speecon test set}} \\
\noalign{\smallskip}\hline\noalign{\smallskip}
Train20            & 22.19         & 21.71          & 22.12         & \textbf{14.78} & 24.54          \\
Train16            & 23.73         & 21.60          & 22.84         & \textbf{14.33} & 24.89          \\
Combined           & 22.43         & 20.42          & 20.93         & \textbf{13.83} & 23.87          \\
\noalign{\smallskip}\hline\noalign{\smallskip}
\multicolumn{6}{c}{\textit{YLE test set}} \\
\noalign{\smallskip}\hline\noalign{\smallskip}
Train20            & 25.41         & 24.89          & 26.15         & \textbf{18.07} & 27.15          \\
Train16            & 27.59         & 26.04          & 27.58         & \textbf{17.61} & 28.37          \\
Combined           & 25.49         & 24.67          & 24.70         & \textbf{17.04} & 26.81          \\
\noalign{\smallskip}\hline
\end{tabular}
\end{table}

For the in-domain test sets, \textit{Train20} and \textit{Combined} acoustic models perform on the same level while \textit{Train16} AM is clearly behind them. The gap is notable especially on the \textit{Test20} set which we believe is due to temporal shift in the data distribution and different segmentation processes. This hypothesis is also supported by \textit{Train16} AM performing best with the Parl-30M LM on \textit{Test20} set, because the Parl-30M LM contains data up to 2020 which the AM is missing.

On the out-of-domain test sets, \textit{Combined} AM performs consistently better than the other AMs with all the LMs, except with the Parl-20M LM. The results for Lahjoita Puhetta test set show how different the colloquial and spontaneous speech in it is from the formal and planned speech in the Finnish parliament. The Conversational LM bridges some of that gap but the WER still stays high when compared to other test sets. Speecon and YLE test sets get best results with general-domain LM and the WERs are lower than for Lahjoita Puhetta. This implies that the domains of Speecon and YLE test sets are much closer to Finnish parliament data than Lahjoita Puhetta.

The thorough evaluation of the HMM-DNN systems establishes a strong baseline against which to compare end-to-end models. The HMM-DNN systems and AED models are compared in an equal data setting: besides choices we used in all HMM-DNN experiments, such as grapheme lexicons, this amounts to limiting the HMM-DNN systems to the transcript-based LMs. Table~\ref{tab:e2e-comparison} lists the results. The corresponding HMM-DNN system outperforms the AED model in every comparison. The AED results follow the same trends as the HMM-DNN systems, with Pearson correlation coefficient value $0.997$ across all the listed results.

\begin{table}
\caption{Equal data comparison of AED models and HMM systems. Results in WER~[\%]. The AED results additionally show the relative WER difference with the corresponding HMM-DNN system in brackets. LP refers to Lahjoita Puhetta test set.}
\label{tab:e2e-comparison}
\centering
\begin{tabular}{lrrr|rrr}
\hline\noalign{\smallskip}
                      & \multicolumn{3}{c}{\textit{HMM/TDNN-medium}} & \multicolumn{3}{c}{\textit{AED}}  \\
\textbf{Test set}     & \textbf{Train20}    & \textbf{Train16}  & \textbf{Comb}    & \textbf{Train20}          & \textbf{Train16}          & \textbf{Comb}              \\
\noalign{\smallskip}\hline\noalign{\smallskip}
Dev16                 & 14.19               & 13.97             & 13.08            & 16.72 (+17.8\%)           & 14.39 (\textbf{+3.01\%})  & 14.09 (+7.72\%)            \\
Test16                & 10.52               & 11.17             & 9.83             & 12.68 (+20.5\%)           & 12.28 (+9.94\%)           & 10.69 (\textbf{+8.75\%})   \\
Test20                & 8.84                & 13.90             & 8.76             & 10.30 (+16.5\%)           & 14.80 (\textbf{+6.47\%})  & 10.15 (+15.9\%)            \\
LP                    & 66.90               & 68.41             & 64.99            & 90.06 (+34.6\%)           & 82.52 (\textbf{+20.6\%})  & 79.78 (+22.8\%)            \\
Speecon               & 22.12               & 22.84             & 20.93            & 25.14 (+13.7\%)           & 25.84 (+13.1\%)           & 21.89 (\textbf{+4.59\%})   \\
YLE                   & 26.15               & 27.58             & 24.70            & 28.99 (\textbf{+10.9\%})  & 31.19 (+13.1\%)           & 29.37 (+18.9\%)            \\
\noalign{\smallskip}\hline
\end{tabular}
\end{table}

\section{Analysis and discussion}
\label{sec:analysis}

We start our discussion with the test set evaluations, their implications, and relation to previous results. Next, we analyse model performance on \textit{Combined} set through the speaker metadata we have and also take a look at the types of substitution errors. We continue our discussion with the comparison of AED and HMM models. Finally, we close discussion by listing possible future research directions.

\subsection{Test set evaluations}

As we expected, in-domain models perform the strongest on in-domain test sets. Yet, it is unexpected how the transcript LM gives the best results on \textit{Test20} set with the \textit{Train20} and \textit{Combined} acoustic models. Since \textit{Test20} set is a product of the same segmentation script as \textit{Train20} set, the sentences in the samples can start and end in the middle. On the other hand, \textit{Train16} set, Parl-20M LM and Parl-30M LM are trained on complete sentences. Therefore, we speculate that models trained on full sentences suffer from a mismatch on the broken sentence structures that appear in the \textit{Test20} set. We think the same phenomenon happens with the general-domain LM that does better than the in-domain Parl-20M LM on the \textit{Test20} set.

The in-domain results also indicate the parliament data distribution shifts with time. In Table~\ref{tab:testset-lm-comparison}, the \textit{Train16} acoustic model performs 4-5 absolute percentage points worse on the \textit{Test20} set than other AMs when LM is in-domain. Furthermore, on the same test set the Parl-30M LM brings improvements for all AMs and the gains are biggest with \textit{Train16} AM.

For the spontaneous speech of Lahjoita Puhetta test set, the conversational LM is the best match. However, the error rates are still more than twice as large compared to models trained on Lahjoita Puhetta data~\citep{moisio2022lahjoita}. This suggests that the colloquial speaking style in Lahjoita Puhetta is acoustically very different from the formal and planned speech in the Parliament data. \textit{Combined} with the general-domain LM, models trained on the parliament data generalise relatively well to the YLE and Speecon data, compared with for instance the results in \citet{mansikkaniemi2017automatic}. Generally, WERs are lower in \cite{mansikkaniemi2017automatic}, where i-vectors and data cleanup were used, and the decreases in WER are in line with gains expected from the aforementioned methods.

An overall trend in the Table~\ref{tab:testset-lm-comparison} shows that for in-domain tasks both the \textit{Train20} and the \textit{Train16} subsets are sufficient alone. Adding more parliament data did not improve models. Conversely, for out-of-domain tasks, more parliament data was consistently better although the gains were small.

\subsection{Error and bias analysis}

Since the 5-hour held out test set is far too small to properly analyse the ASR errors, we create another setup by taking only 100 hours of speech for training and leaving most of the original training data for a huge held-out test set. We make sure there is no overlap in speakers between the two sets. Here we assume that the models trained with randomly sampled 100 hours would already be good enough to indicate the speech which is most difficult to recognise.

Specifically, we train a TDNN-medium model using a dataset to which random speakers' utterances were added until a total of 40\,000 speech samples was reached - this corresponds to roughly 100 hours of training data. This 100h subset ended up having slightly less speech from women and older speakers compared to the full data. We then decode the remaining \textit{Combined} set with this TDNN-medium-100h model and the general-domain LM. We chose the general-domain LM to avoid using the transcripts and other material related to the chosen test data and because a transcript LM of only 100 hours would make a poor LM.

\begin{figure}[hp]
    \centering
    \begin{subfigure}[b]{0.47\textwidth}
        \includegraphics[width=\textwidth]{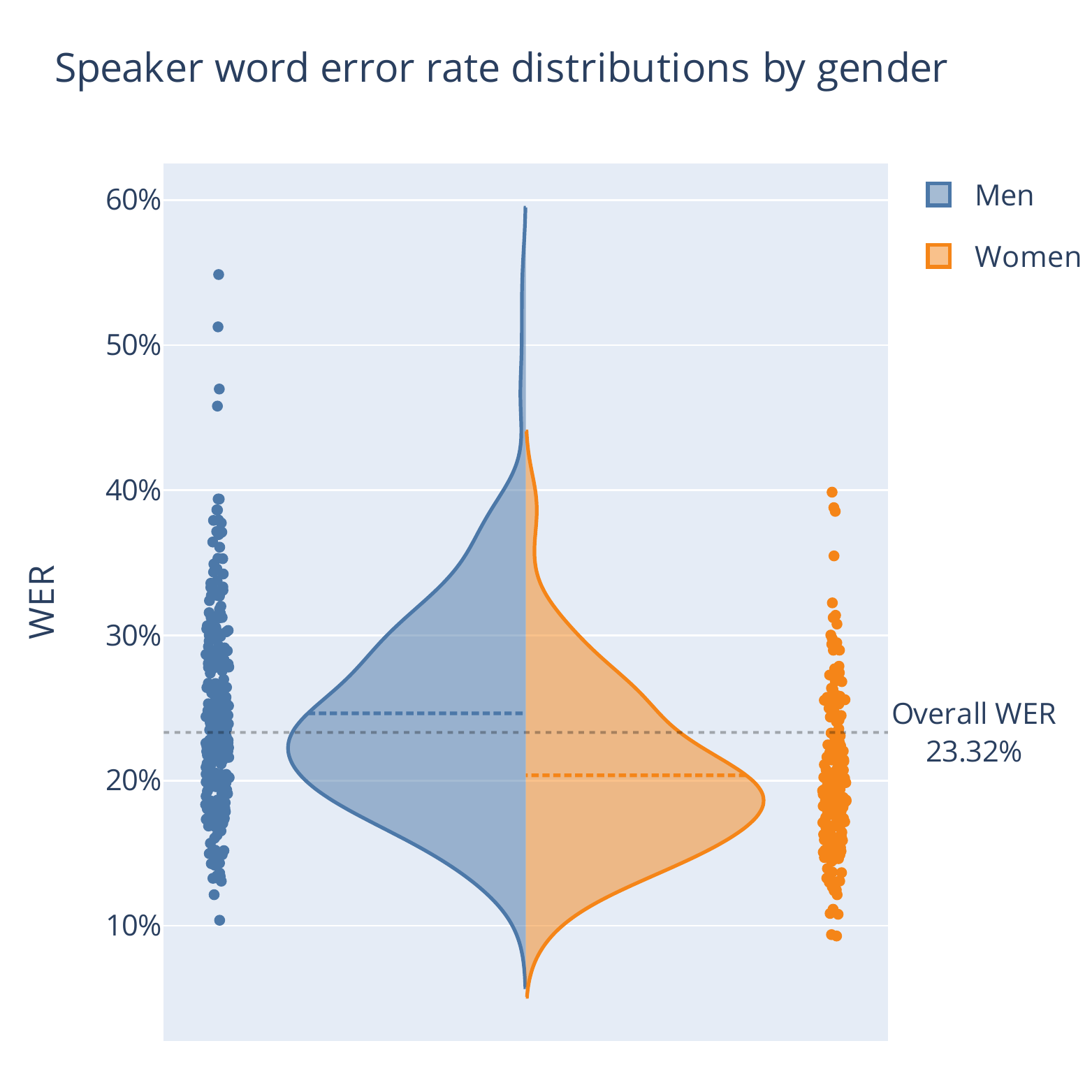}
        \caption{Women have lower WER on average.}
        \label{fig:wer_gender}
    \end{subfigure}
    \begin{subfigure}[b]{0.47\textwidth}
        \includegraphics[width=\textwidth]{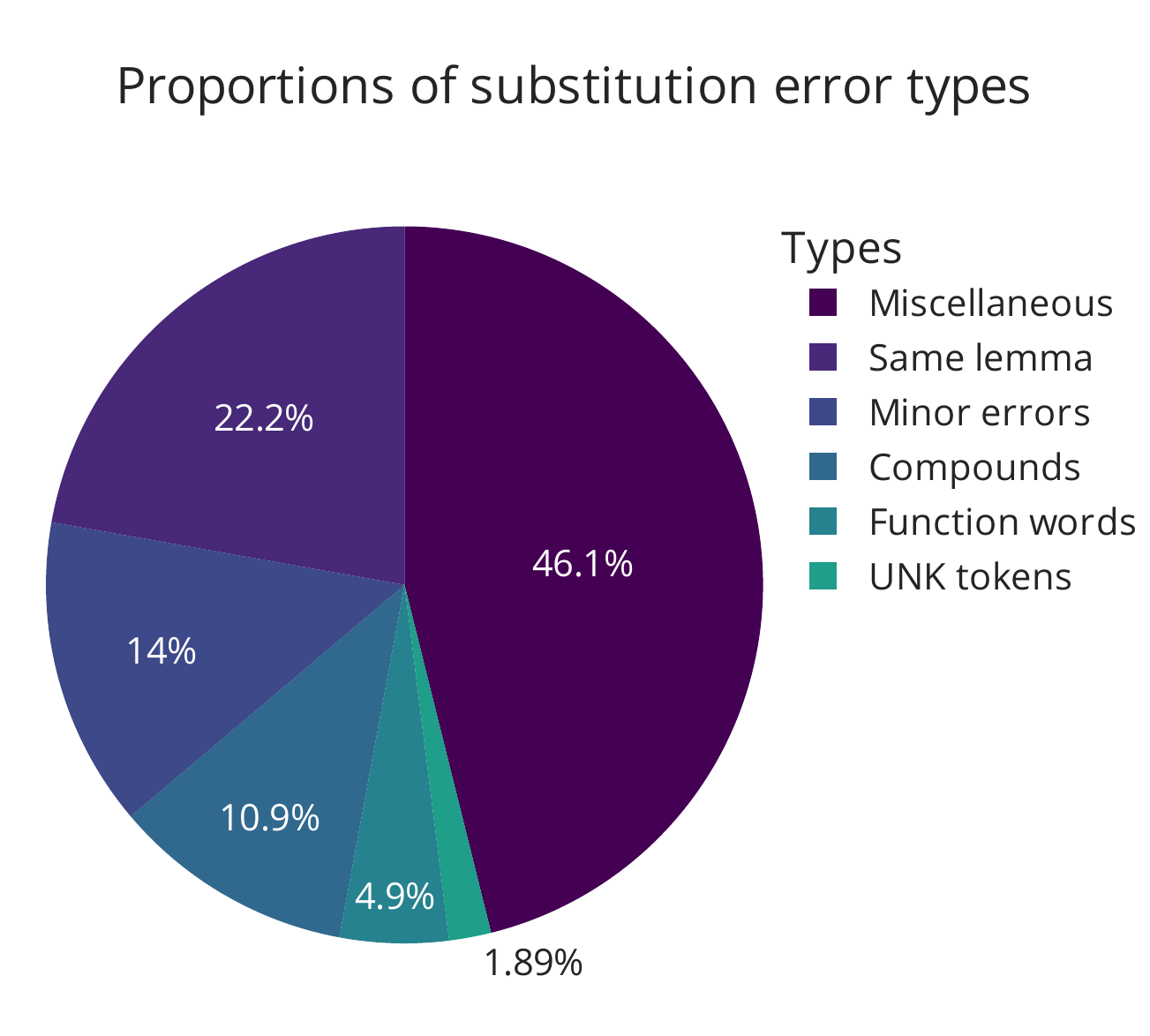}
        \caption{Over half of substitutions can be categorised.}
        \label{fig:sub_error_pie}
    \end{subfigure}
    \hfill
    \begin{subfigure}[b]{0.99\textwidth}
        \includegraphics[width=\textwidth]{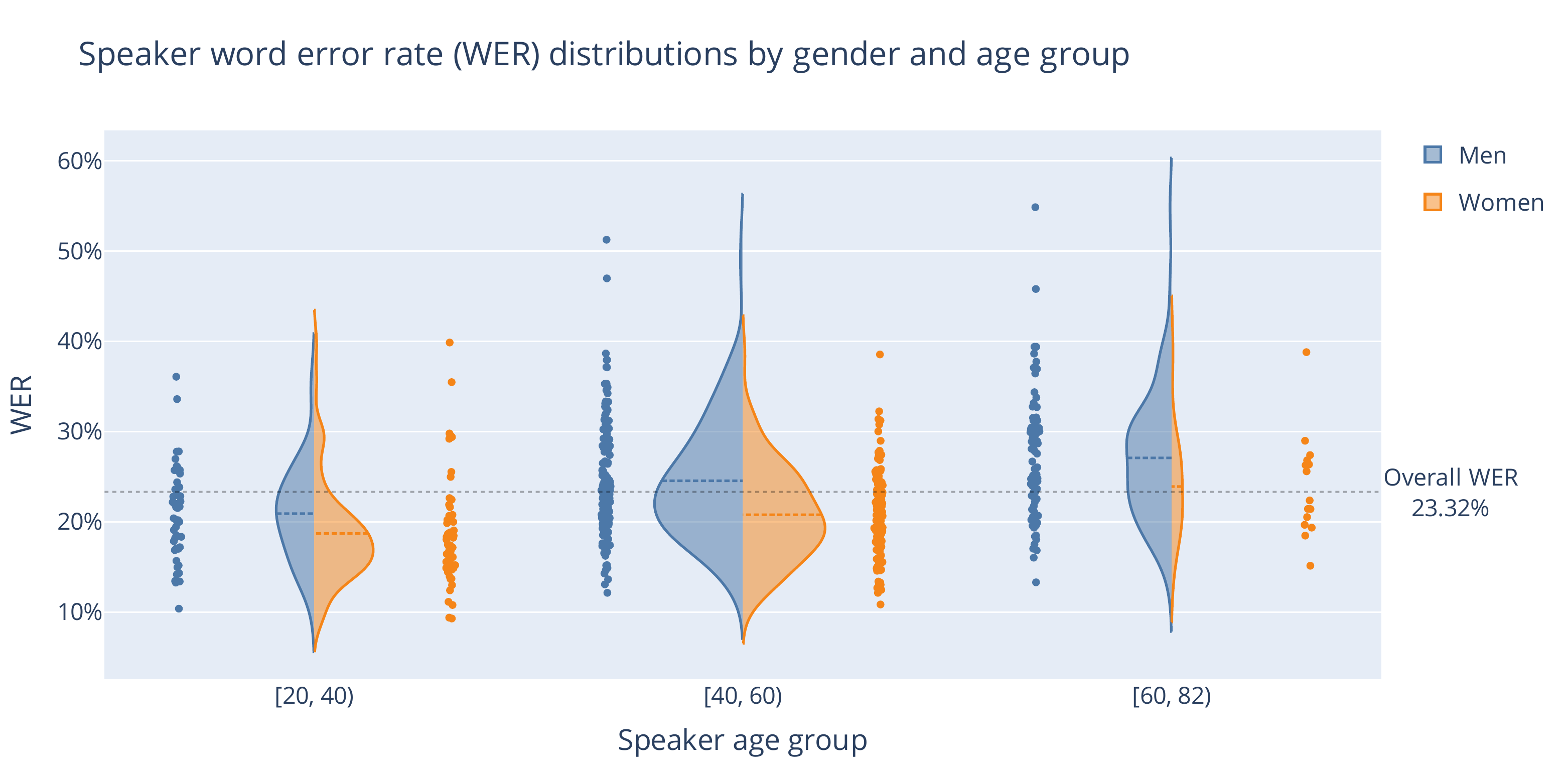}
        \caption{WER is on average higher for older speakers of both genders.}
        \label{fig:wer_age_group}
    \end{subfigure}
    \hfill
    \begin{subfigure}[b]{0.9\textwidth}
        \includegraphics[width=\textwidth]{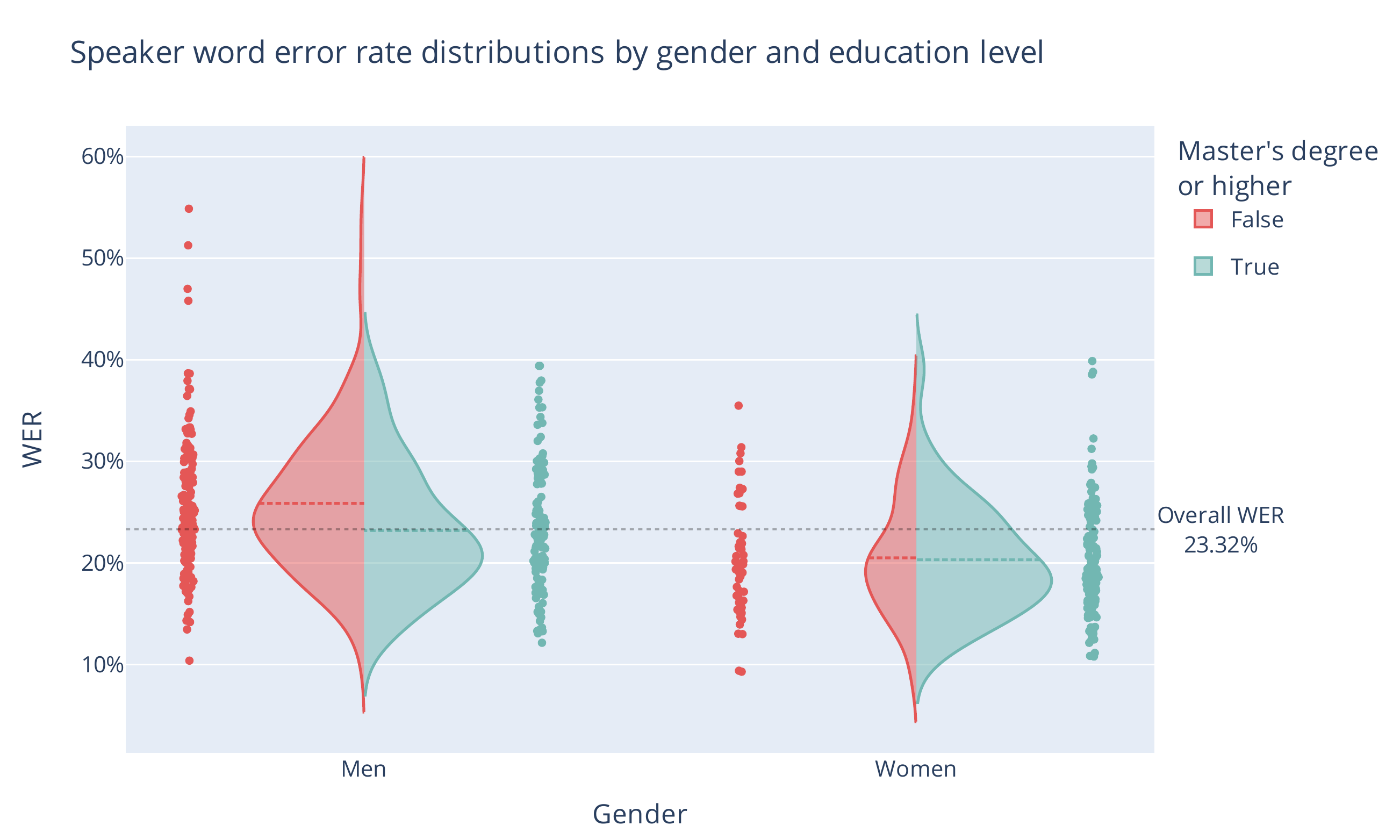}
        \caption{WER is on average lower for men with at least Master's degree.}
        \label{fig:wer_education_gender}
    \end{subfigure}
    \caption{This figure shows closer looks at the gender, speaker age, and education level distributions as well as what kind of substitution errors the TDNN-medium-100h makes. Since a speaker may have contributed samples over many years, we calculate an average age for each speaker from their samples.}
    \label{fig:analysis_plots}
\end{figure}

It is not surprising that the model trained with 100 hours is worse than the one trained with over 3000 hours, e.g. 22.64~\% versus 14.52~\% on \textit{Dev16}, with the general-domain LM. For the full decoded set that we use in the analysis the WER is 23.32~\%. In Figures \ref{fig:wer_gender}, \ref{fig:wer_age_group}, and \ref{fig:wer_education_gender} we visualise how this WER result is distributed among speakers of different gender, age groups and education level. In the age plot we have calculated an average age for each speaker from all of their samples, because most speakers have contributed samples over several years. Between genders it is clear that women's speech is easier to recognise in this corpus despite women's utterances only making up 22~\% of the 100h subset. This difference between genders also persists between the three age groups and education levels. Age-wise, the speech of younger speakers is easier to recognise than that of older ones. However, that observation may be partly explained by the skew toward younger speakers in the 100h subset. Education level on the other hand, appears to matter only for men. We estimated the effects of speaker dialect (assumed based on speaker birthplace), and speaker political party, but these variables did not have a statistically significant effect on WER in a multivariate ordinary least squares model.

In addition to the speaker distributions, we studied the types of substitution errors the TDNN-medium-100h model makes. We were able to categorise over half of the substitution errors into recognisable types, as is shown in Figure \ref{fig:sub_error_pie}. In 22.2~\% of the errors the reference word and ASR hypothesis have the same lemma. For an agglutinative language like Finnish, it is reasonable that there are many errors related to affixes and inflections. Minor errors appear in 14~\% of the cases. We consider substitution error as minor when reference and hypothesis words have different lemmas and edit distance is, at maximum, either one for words up to four characters or two for longer words. 10.9~\% of the errors are related to compound words; either separate words are incorrectly compounded or compounded words separated in the hypothesis. Despite being a minor error, compound mistakes have a large effect on WER because they imply another insertion or deletion error as well. Function word errors, 4.9~\%, occur when one function word is substituted with another. We think many of these are related to the clarity and readability edits parliament transcribers make in the transcripts. The few UNK errors are caused by the UNK tokens Kaldi segmentation script adds to the segment transcripts to mark unknown acoustics.

\subsection{AED versus HMM}

End-to-end AED models can be trained on the \textit{Train20}, \textit{Train16}, and the \textit{Combined} datasets without any pre-processing steps. We took care to create a balanced comparison between the AED models and HMM systems: in Table~\ref{tab:e2e-comparison}, both models used the same data (just transcribed speech), neither model leveraged augmentation/i-vectors, and the total learned parameter counts were about the same (27.7M for AED vs. 18.8M in HMM acoustic model + 10M parameters in the n-gram LM). 

The AED models consistently trail the corresponding HMM systems in WER. Although in absolute numbers the AED models fare worse, relatively they seem to generalise as well as the HMM systems, except for the Lahjoita Puhetta data, where the AED models essentially fail. The HMM system results in table~\ref{tab:lm-comparison} suggest that for in-domain acoustic modeling, the 2020 or the 2016 datasets are sufficient alone, so we hypothesise that most of AED model or transcript LM HMM system improvements gained on the \textit{Combined} dataset result primarily from the model seeing more varied text.

\subsection{Future work}

The HMM system recipe provides a tuned HMM-GMM system, which can serve to speed up neural network acoustic model research. Furthermore, the full recipe serves as a benchmark. Future work in neural language model rescoring could improve both in-domain as well as out-of-domain results.

For AEDs our recipe provides a starting point and benchmark for future research, where Transformer architectures and other larger models could possibly improve the results and possibly even close the performance gap with HMM systems. Additionally, shallow fusion neural language models trained on the 30M token corpus would be a worthwhile experiment, although the resulting system is no longer trained end-to-end.

This corpus has a relatively rich set of metadata, making it suitable for many speech classification/diarization tasks, such as speaker recognition. A rare possibility is a longitudinal study of how the representatives' speech changes over time. The visual stream in the parliament video recordings provides a way to study audio-visual and multimodal speech recognition for Finnish. To further enrich the corpus, public meeting data can also be collected from other sources. For instance, some city and local councils in Finland record their meetings and make them available online\footnote{See for example (only available in Finnish/Swedish) \url{https://www.raasepori.fi/etusivu/paatoksenteko/esityslistat-ja-poytakirjat/}}. Even if there would be no meeting transcripts available, the data could be harnessed for unsupervised training.

\section{Conclusions}
\label{sec:conclusions}

In this paper, we have presented a new, extended version of the Finnish parliament ASR corpus. Totalling at over 3 000 hours of data, it is the largest transcribed Finnish ASR corpus we know. The corpus has two official temporally distinct subsets, and two temporally distinct, manually corrected test sets, as well as a development set. We have developed benchmark models for three ASR approaches - HMM-GMM, HMM-DNN, and AED. 

Our optimised HMM-GMM recipe can be leveraged to kick-start new research. The HMM-DNN and AED recipes provide starting points and benchmarks. Despite the large-scale of the data, the HMM-DNN approach consistently outperforms the AED approach when compared in a matched equal data setting in our experiments. The experiments show that this dataset is suitable for training ASR systems for many types of planned or formal Finnish, but our models do not generalise to colloquial speech.

The rich metadata allowed us to analyse the errors our models make. Women, younger representatives, and at least Master's level education attainees have lower word error rates than their counterparts. 

New parliament sessions can be processed with our new retrieval and segmentation pipeline as it becomes available. Our results suggest that on the temporally older test set performance already plateaus at the current scale. However, models trained on the older training subset do not perform as well on the newer test set, suggesting that new data is still necessary to keep up with some shifting data characteristics.

\begin{acknowledgements}
This work has been supported by the MeMAD project of the European Union’s Horizon 2020 research and innovation programme (grant agreement No 780069) and the Academy of Finland project funding grants number 329267, 337073, and 345790. We also thank Aalto ScienceIT for providing us computational resources.
\end{acknowledgements}

\section*{Statements and Declarations}
The authors have no relevant financial or non-financial interests to disclose.

\bibliographystyle{spbasic}      % basic style, author-year citations
\bibliography{refs.bib}   % name your BibTeX data base

\begin{thebibliography}{32}
\providecommand{\natexlab}[1]{#1}
\providecommand{\url}[1]{{#1}}
\providecommand{\urlprefix}{URL }
\expandafter\ifx\csname urlstyle\endcsname\relax
  \providecommand{\doi}[1]{DOI~\discretionary{}{}{}#1}\else
  \providecommand{\doi}{DOI~\discretionary{}{}{}\begingroup
  \urlstyle{rm}\Url}\fi
\providecommand{\eprint}[2][]{\url{#2}}

\bibitem[{{Bahdanau} et~al(2016){Bahdanau}, {Chorowski}, {Serdyuk}, {Brakel},
  and {Bengio}}]{bahdanau2016attention}
{Bahdanau} D, {Chorowski} J, {Serdyuk} D, {Brakel} P, {Bengio} Y (2016)
  End-to-end attention-based large vocabulary speech recognition. In:
  \emph{2016 IEEE International Conference on Acoustics, Speech and Signal
  Processing (ICASSP)}, pp 4945--4949, \doi{10.1109/ICASSP.2016.7472618}

\bibitem[{{Chan} et~al(2016){Chan}, {Jaitly}, {Le}, and
  {Vinyals}}]{chan2016las}
{Chan} W, {Jaitly} N, {Le} Q, {Vinyals} O (2016) Listen, attend and spell: A
  neural network for large vocabulary conversational speech recognition. In:
  \emph{2016 IEEE International Conference on Acoustics, Speech and Signal
  Processing (ICASSP)}, pp 4960--4964, \doi{10.1109/ICASSP.2016.7472621}

\bibitem[{Chiu et~al(2019)Chiu, Han, Zhang, Pang, Kishchenko, Nguyen,
  Narayanan, Liao, Zhang, Kannan, Prabhavalkar, Chen, Sainath, and
  Wu}]{chiu2019long-form}
Chiu CC, Han W, Zhang Y, Pang R, Kishchenko S, Nguyen P, Narayanan A, Liao H,
  Zhang S, Kannan A, Prabhavalkar R, Chen Z, Sainath T, Wu Y (2019) A
  comparison of end-to-end models for long-form speech recognition. In:
  \emph{2019 IEEE Automatic Speech Recognition and Understanding Workshop
  (ASRU)}, pp 889--896, \doi{10.1109/ASRU46091.2019.9003854}

\bibitem[{Enarvi(2018)}]{enarvi2018}
Enarvi S (2018) {Modeling Conversational Finnish for Automatic Speech
  Recognition}. PhD thesis, Aalto University School of Electrical Engineering

\bibitem[{Enarvi et~al(2017)Enarvi, Smit, Virpioja, and
  Kurimo}]{enarvi2017convfinnish}
Enarvi S, Smit P, Virpioja S, Kurimo M (2017) {Automatic speech recognition
  with very large conversational Finnish and Estonian vocabularies}.
  \emph{IEEE/ACM Transactions on Audio, Speech, and Language Processing}
  25(11):2085--2097, \doi{10.1109/TASLP.2017.2743344}

\bibitem[{Geneva et~al(2019)Geneva, Shopov, and Mihov}]{geneva2019building}
Geneva D, Shopov G, Mihov S (2019) Building an {ASR corpus based} on {Bulgarian
  Parliament speeches}. In: \emph{Statistical {Language} and {Speech
  Processing}}, Springer, pp 188--197

\bibitem[{Helgadóttir et~al(2017)Helgadóttir, Kjaran, Nikulásdóttir, and
  Guðnason}]{helgadottir2017building}
Helgadóttir IR, Kjaran R, Nikulásdóttir AB, Guðnason J (2017) {Building an
  ASR corpus Using Althingi’s Parliamentary speeches}. In: \emph{Proc.
  Interspeech 2017}, pp 2163--2167, \doi{10.21437/Interspeech.2017-903}

\bibitem[{Höge et~al(1999)Höge, Draxler, van~den Heuvel, Johansen, Sanders,
  and Tropf}]{hoge1999speechdat}
Höge H, Draxler C, van~den Heuvel H, Johansen FT, Sanders E, Tropf HS (1999)
  {Speechdat multilingual speech databases for teleservices: across the finish
  line}. In: \emph{Proc. 6th European Conference on Speech Communication and
  Technology (Eurospeech 1999)}, pp 2699--2702

\bibitem[{Imseng et~al(2012)Imseng, Bourlard, Caesar, Garner, Lecorvé, and
  Nanchen}]{imseng2012mediaparl}
Imseng D, Bourlard H, Caesar H, Garner PN, Lecorvé G, Nanchen A (2012)
  {MediaParl}: Bilingual mixed language accented speech database. In:
  \emph{2012 IEEE Spoken Language Technology Workshop (SLT)}, pp 263--268,
  \doi{10.1109/SLT.2012.6424233}

\bibitem[{Iskra et~al(2002)Iskra, Grosskopf, Marasek, van~den Heuvel, Diehl,
  and Kiessling}]{iskra-2002-speecon}
Iskra D, Grosskopf B, Marasek K, van~den Heuvel H, Diehl F, Kiessling A (2002)
  {SPEECON} {--} speech databases for consumer devices: Database specification
  and validation. In: \emph{Proceedings of the Third International Conference
  on Language Resources and Evaluation ({LREC}{'}02)}, European Language
  Resources Association (ELRA), Las Palmas, Canary Islands - Spain, pp 329--333

\bibitem[{Joulin et~al(2017)Joulin, Grave, Bojanowski, and
  Mikolov}]{joulin2017bag}
Joulin A, Grave E, Bojanowski P, Mikolov T (2017) Bag of tricks for efficient
  text classification. In: \emph{Proceedings of the 15th Conference of the
  {E}uropean Chapter of the Association for Computational Linguistics: Volume
  2, Short Papers}, Association for Computational Linguistics, Valencia, Spain,
  pp 427--431, \urlprefix\url{https://aclanthology.org/E17-2068}

\bibitem[{Keung et~al(2020)Keung, Niu, Lu, Salazar, and
  Bhardwaj}]{keung2020attentional}
Keung P, Niu W, Lu Y, Salazar J, Bhardwaj V (2020) Attentional speech
  recognition models misbehave on out-of-domain utterances. \emph{arXiv
  preprint arXiv:200205150}

\bibitem[{{Kim} et~al(2017){Kim}, {Hori}, and {Watanabe}}]{jointattentionctc}
{Kim} S, {Hori} T, {Watanabe} S (2017) Joint {CTC}-attention based end-to-end
  speech recognition using multi-task learning. In: \emph{2017 IEEE
  International Conference on Acoustics, Speech and Signal Processing
  (ICASSP)}, pp 4835--4839, \doi{10.1109/ICASSP.2017.7953075}

\bibitem[{Kirkedal et~al(2020)Kirkedal, Stepanović, and
  Plank}]{kirkedal2020ftspeech}
Kirkedal A, Stepanović M, Plank B (2020) {FT Speech: Danish Parliament Speech
  Corpus}. In: \emph{Proc. Interspeech 2020}, pp 442--446,
  \doi{10.21437/Interspeech.2020-3164}

\bibitem[{Kratochvil et~al(2020)Kratochvil, Polak, and
  Bojar}]{kratochvil-2020-large}
Kratochvil J, Polak P, Bojar O (2020) Large corpus of {C}zech parliament
  plenary hearings. In: \emph{Proceedings of the 12th Language Resources and
  Evaluation Conference}, European Language Resources Association, Marseille,
  France, pp 6363--6367

\bibitem[{Kudo and Richardson(2018)}]{kudo-richardson-2018-sentencepiece}
Kudo T, Richardson J (2018) {S}entence{P}iece: A simple and language
  independent subword tokenizer and detokenizer for neural text processing. In:
  \emph{Proceedings of the 2018 Conference on Empirical Methods in Natural
  Language Processing: System Demonstrations}, Association for Computational
  Linguistics, Brussels, Belgium, pp 66--71, \doi{10.18653/v1/D18-2012}

\bibitem[{Lennes(2009)}]{lennes2009finintas}
Lennes M (2009) {Segmental features in spontaneous and read-aloud Finnish}. In:
  {Viola de Silva, Riikka Ullakonoja} (ed) \emph{Phonetics of Russian and
  Finnish}, Peter Lang, Germany, pp 145--166

\bibitem[{Lindén et~al(2022)Lindén, Jauhiainen, Lennes, Kurimo, Rossi, Kurki,
  and Pitkänen}]{lindenCLARINbook}
Lindén K, Jauhiainen T, Lennes M, Kurimo M, Rossi A, Kurki T, Pitkänen O
  (2022) {Donate Speech}: Collecting and sharing a large-scale speech database
  for {S}ocial {S}ciences, {H}umanities and {A}rtificial {I}ntelligence
  research and innovation. In: Witt A, Fisher D (eds) \emph{CLARIN Book},
  DeGruyter, chap 3.6

\bibitem[{Manohar et~al(2017)Manohar, Povey, and Khudanpur}]{manohar2017jhuMGB}
Manohar V, Povey D, Khudanpur S (2017) {JHU Kaldi system for Arabic MGB-3 ASR
  challenge using diarization, audio-transcript alignment and transfer
  learning}. In: \emph{2017 IEEE Automatic Speech Recognition and Understanding
  Workshop (ASRU)}, pp 346--352, \doi{10.1109/ASRU.2017.8268956}

\bibitem[{Mansikkaniemi et~al(2017)Mansikkaniemi, Smit, and
  Kurimo}]{mansikkaniemi2017automatic}
Mansikkaniemi A, Smit P, Kurimo M (2017) {Automatic construction of the Finnish
  Parliament speech corpus}. In: \emph{Proceedings of the Annual Conference of
  the International Speech Communication Association, INTERSPEECH},
  International Speech Communication Association, pp 3762--3766,
  \doi{10.21437/Interspeech.2017-1115}

\bibitem[{Meyer and Schramm(2006)}]{meyer2006boostingHMM}
Meyer C, Schramm H (2006) Boosting {HMM} acoustic models in large vocabulary
  speech recognition. \emph{Speech Communication} 48(5):532--548,
  \doi{10.1016/j.specom.2005.09.009}

\bibitem[{Moisio et~al(2022)Moisio, Porjazovski, Rouhe, Getman, Virkkunen,
  Gr{\'o}sz, Lind{\'e}n, and Kurimo}]{moisio2022lahjoita}
Moisio A, Porjazovski D, Rouhe A, Getman Y, Virkkunen A, Gr{\'o}sz T,
  Lind{\'e}n K, Kurimo M (2022) {Lahjoita puhetta--a large-scale corpus of
  spoken Finnish with some benchmarks}. \emph{arXiv preprint arXiv:220312906}

\bibitem[{Pl{\"{u}}ss et~al(2020)Pl{\"{u}}ss, Neukom, and
  Vogel}]{pluss2020swiss}
Pl{\"{u}}ss M, Neukom L, Vogel M (2020) {Swiss Parliaments corpus, an
  automatically aligned Swiss German speech to standard German text corpus}.
  \emph{arXiv preprint arXiv:201002810}

\bibitem[{Povey et~al(2011)Povey, Ghoshal, Boulianne, Burget, Glembek, Goel,
  Hannemann, Motlicek, Qian, Schwarz, Silovsky, Stemmer, and
  Vesely}]{povey2011kaldi}
Povey D, Ghoshal A, Boulianne G, Burget L, Glembek O, Goel N, Hannemann M,
  Motlicek P, Qian Y, Schwarz P, Silovsky J, Stemmer G, Vesely K (2011) {The
  Kaldi Speech Recognition Toolkit}. In: \emph{IEEE 2011 Workshop on Automatic
  Speech Recognition and Understanding}, IEEE Signal Processing Society

\bibitem[{Povey et~al(2016)Povey, Peddinti, Galvez, Ghahremani, Manohar, Na,
  Wang, and Khudanpur}]{Povey2016lfmmi}
Povey D, Peddinti V, Galvez D, Ghahremani P, Manohar V, Na X, Wang Y, Khudanpur
  S (2016) Purely sequence-trained neural networks for {ASR} based on
  {Lattice-Free MMI}. In: \emph{Interspeech 2016}, pp 2751--2755,
  \doi{10.21437/Interspeech.2016-595}

\bibitem[{Povey et~al(2018)Povey, Cheng, Wang, Li, Xu, Yarmohammadi, and
  Khudanpur}]{povey18_interspeech}
Povey D, Cheng G, Wang Y, Li K, Xu H, Yarmohammadi M, Khudanpur S (2018)
  {Semi-Orthogonal Low-Rank Matrix Factorization for Deep Neural Networks}. In:
  \emph{Proc. Interspeech 2018}, pp 3743--3747,
  \doi{10.21437/Interspeech.2018-1417}

\bibitem[{Ravanelli et~al(2021)Ravanelli, Parcollet, Plantinga, Rouhe, Cornell,
  Lugosch, Subakan, Dawalatabad, Heba, Zhong, Chou, Yeh, Fu, Liao, Rastorgueva,
  Grondin, Aris, Na, Gao, Mori, and Bengio}]{speechbrain}
Ravanelli M, Parcollet T, Plantinga P, Rouhe A, Cornell S, Lugosch L, Subakan
  C, Dawalatabad N, Heba A, Zhong J, Chou JC, Yeh SL, Fu SW, Liao CF,
  Rastorgueva E, Grondin F, Aris W, Na H, Gao Y, Mori RD, Bengio Y (2021)
  Speechbrain: A general-purpose speech toolkit. \eprint{2106.04624}

\bibitem[{Rouhe et~al(2021)Rouhe, Van~Camp, Singh, Van~Hamme, and
  Kurimo}]{rouhe2021equal}
Rouhe A, Van~Camp A, Singh M, Van~Hamme H, Kurimo M (2021) An equal data
  setting for attention-based encoder-decoder and {HMM/DNN} models: A case
  study in {Finnish ASR}. In: Karpov A, Potapova R (eds) \emph{Speech and
  Computer}, Springer International Publishing, Cham, pp 602--613

\bibitem[{Sennrich et~al(2016)Sennrich, Haddow, and Birch}]{sennrich2016bpe}
Sennrich R, Haddow B, Birch A (2016) Neural machine translation of rare words
  with subword units. In: \emph{Proceedings of the 54th Annual Meeting of the
  Association for Computational Linguistics (Volume 1: Long Papers)},
  Association for Computational Linguistics, Berlin, Germany, pp 1715--1725,
  \doi{10.18653/v1/P16-1162}

\bibitem[{Siivola et~al(2007)Siivola, Creutz, and Kurimo}]{siivola2007varikn}
Siivola V, Creutz M, Kurimo M (2007) {Morfessor and VariKN machine learning
  tools for speech and language technology}. In: \emph{8th Annual Conference of
  the International Speech Communication Association (Interspeech 2007),
  Antwerp, Belgium, August 27-31, 2007}, ISCA, pp 1549--1552

\bibitem[{Voutilainen(2017)}]{voutilainen2017fiparl}
Voutilainen E (2017) {The regulation of linguistic quality in the official
  speech-to-text reports of the Finnish Parliament}. \emph{CoMe: Studies on
  Communication and Linguistic and Cultural Mediation} p 61–73

\bibitem[{Wang et~al(2021)Wang, Riviere, Lee, Wu, Talnikar, Haziza, Williamson,
  Pino, and Dupoux}]{wang-2021-voxpopuli}
Wang C, Riviere M, Lee A, Wu A, Talnikar C, Haziza D, Williamson M, Pino J,
  Dupoux E (2021) {V}ox{P}opuli: {A large-scale multilingual speech corpus for
  representation learning, semi-supervised learning and interpretation}. In:
  \emph{Proceedings of the 59th Annual Meeting of the Association for
  Computational Linguistics and the 11th International Joint Conference on
  Natural Language Processing (Volume 1: Long Papers)}, Association for
  Computational Linguistics, Online, pp 993--1003,
  \doi{10.18653/v1/2021.acl-long.80}

\end{thebibliography}

\end{document}